\documentclass[twoside,11pt]{article}

\usepackage[abbrvbib, preprint]{jmlr2e}

\usepackage[inline]{enumitem}
\usepackage{multirow}

\ShortHeadings{Frouros: A Python library for drift detection in machine learning systems}{Céspedes Sisniega and López García}
\firstpageno{1}

\begin{document}

\title{Frouros: A Python library for drift detection in machine learning systems}

\author{\name Jaime Céspedes Sisniega \email cespedes@ifca.unican.es \\
       \name Álvaro López García \email aloga@ifca.unican.es \\
       \addr Instituto de Física de Cantabria (IFCA), CSIC-UC, Spain
}
\editor{}

\maketitle

\begin{abstract}
    \verb+Frouros+ is an open-source Python library capable of detecting drift in machine learning systems. It provides a combination of classical and more recent algorithms for drift detection: both concept and data drift. We have designed it with the objective of making it compatible with any machine learning framework and easily adaptable to real-world use cases. The library is developed following a set of best development and continuous integration practices to ensure ease of maintenance and extensibility. The source code is available at \url{https://github.com/IFCA/frouros}.
\end{abstract}

\begin{keywords}
  Machine learning, Drift detection, Concept drift, Data drift, Python
\end{keywords}

\section{Introduction}

When deploying machine learning models in real-world applications, there is often the erroneous assumption that a given model will be used in a stationary environment, assuming that the same concepts learned during the training phase will remain valid at inference time \citep{gama2004learning}, or that training samples and production-time samples will come from the same distribution \citep{ackerman2021automatically}. However, in real-world scenarios, this is often far from being always true, and both situations may result in some type of \textit{drift} that can end up affecting the model performance \citep{vzliobaite2016overview}. Additionally, due to the high cost of collecting and labeling samples, this performance loss can often not be confirmed in many real-world problems, and other methods that only rely on distribution changes must be used.

In this paper, we present \verb+Frouros+, an open-source Python library for drift detection in machine learning systems. The library tries to fulfill two main objectives:
\begin{enumerate*} 
    \item to easily integrate into a machine learning system workflow that uses any machine learning framework, thus making it framework-agnostic;
    \item to unify in a single library the \textit{concept drift} detection part (traditionally researched and used for streaming/evolving data streams and incremental learning as described by \citealp{khamassi2018discussion}) with the research of change detection in the covariate distributions (also known as \textit{dataset shift} or \textit{distribution shift}, related to the field of statistical two-sample testing as introduced in \citealp{rabanser2019failing}, and methods that measure the distance between distributions as described by \citealp{goldenberg2019survey}).
\end{enumerate*}

\section{Drift detection}
\label{sec:drift-detection}

Traditionally there has been little consensus on the terminology and definitions of the different types of \textit{drift}, as stated by \cite{moreno2012unifying}. In order to adopt some clear definitions for the remainder of this paper, we apply those used by \cite{gama2014survey} for the \textit{concept drift} part, in combination with those used by \cite{rabanser2019failing} work for detecting \textit{dataset shift} using only the covariates. Therefore, we set up the following definitions assuming two different time points, ${t}$ and ${t + w}$, where $t$ could be any point in time, and $w$ could be the time at which the existence of change is checked \citep{bayram2022concept}. Thus, given $P(X, y) = P(y|X)P(X)$, a change in the joint distribution between two different times that can result in performance degradation can be described as $P_{t}(X, y) \neq P_{t + w}(X, y)$. The types of changes that can lead to degradation of the model's performance are categorized as follows:

\paragraph{Concept drift.} There is a change in the conditional probability $P(y|X)$, with or without a change in $P(X)$. Thus, it can be defined as $P_{t}(y|X) \neq P_{t + w}(y|X)$. Also known as \textit{real concept drift} \citep{gama2014survey}.

\paragraph{Data drift.} There is a change in $P(X)$. Therefore, this type of drift only focuses in the distribution of the covariates, $P(X)$, so $P_{t}(X) \neq P_{t + w}(X)$. Hereafter, we rename \textit{dataset shift} \citep{rabanser2019failing} to \textit{data drift}, in order to maintain consistency with the above definitions and with some of the related software mentioned in Section \ref{sec:comparison} that also refer to it as \textit{data drift} \citep{alibi-detect, menelaus}. A variation of \textit{data drift} can be considered \textit{label drift}, also known as \textit{prior probability shift} \citep{storkey2009training}, where the change occurs in the label distribution $P(Y)$ instead of the covariates $P(X)$. Thus, $P_{t}(Y) \neq P_{t + w}(Y)$.

\section{Overview and design}
\label{sec:overview-design}

The design and implementation of the library have been carried out with the aim of making it compatible with any machine learning framework, thus making it framework-agnostic. Detection methods are divided into the \textit{concept drift} and \textit{data drift} categories, depending on the type of drift they can detect, according to the definitions given in Section~\ref{sec:drift-detection}, and how they detect it. In terms of implementation, both categories share an abstract class named \verb+BaseDetector+ that contains the basic methods (reset) and attributes (callbacks) that a detectors must implement. \verb+Frouros+ also implements the well-known concept of \textit{callbacks}, presented in some machine learning frameworks, such as \verb+Keras+ \citep{chollet2015keras} and \verb+PyTorch Lightning+ \citep{Falcon_PyTorch_Lightning_2019}. This allows users to execute custom code at certain points. Additionally, we provide prequential error metrics \citep{gama2013evaluating} to evaluate the performance of \textit{concept drift} methods described in Section~\ref{subsec:concept-drift}.

\subsection{Concept drift}
\label{subsec:concept-drift}

These methods are aimed at detecting \textit{concept drift}. In order to update the detector, depending on the type of algorithm, they may require either the ground-truth labels/values or the values of the variable to be tracked, which is usually a model performance metric. The detector can receive the update values on demand; therefore, it is not mandatory to feed it with all the ground-truth labels/values or model performance metric values. All implemented methods work in a streaming manner, receiving one sample at a time.

In terms of implementation, each detector inherits from the \verb+BaseConceptDrift+ abstract class. Depending on the type of \textit{concept drift} detector, it also inherits from its corresponding type abstract class. The \verb+update+ method is used to update the detector's inner state, and it checks if drift is occurring each time a new value is received. Additionally, upon instantiation, the detector can optionally receive a list of \textit{callbacks}. It also receives a configuration class that contains specific parameters determining its behavior.

\subsection{Data drift}
\label{subsec:data-drift}

\textit{Data drift} methods focus on detecting changes by considering only the covariates. Therefore, these algorithms try to detect changes at a feature level by comparing new data distributions against reference data distributions.

Depending on the number of samples given to the detector to compare with the reference distribution, these methods can be classified as batch or streaming. The former uses a batch of samples to test against the reference distribution, and the latter can receive one sample at a time to update the inner state of the detector, therefore the samples must follow a sequential order.

Unlike \textit{concept drift} detectors, there is a \verb+fit+ method that stores the samples from the reference distribution used for comparison at inference time.
In the case of batch algorithms, the \verb+compare+ method is used to carry out the comparison and obtain a statistic or distance value depending on the type of detector. Whereas for the streaming detectors, an \verb+update+ method is used each time a new sample is added to the detector.

Based on the type of \textit{data drift} detector (statistical test or distance-based), checking if drift is taking place is done in different ways. After calling the \verb+compare+ or \verb+update+ methods, statistical tests provide the corresponding p-value that determines if drift exists. For distance-based detectors, the output is only the distance between the provided data and the data from the reference distribution. Therefore through the use of \textit{callbacks}, explained in Section \ref{subsec:callbacks}, p-values can be obtained. 

Moreover, \textit{data drift} detectors are implemented considering the type of data that they are expected to work with (categorical or numerical) and the number of features that can be considered (univariate or multivariate).

As mentioned in Section \ref{sec:drift-detection}, \textit{data drift} methods can also be used for detecting \textit{label drift} by considering the label distribution. Consequently, the described way of using the detectors for \textit{data drift} applies to \textit{label drift}.

\subsection{Callbacks}
\label{subsec:callbacks}

Execution of custom code at certain points is possible through the use of \textit{callbacks}. This adds enough flexibility to \verb+Frouros+ to be adapted to real-world use cases required by the users. \textit{Callbacks} are divided according to whether the detector belongs to the batch or streaming category.

For batch detectors, \textit{callbacks} inherit from \verb+BaseCallbackBatch+ class, which provides methods such as \verb+on_compare_start+ and \verb+on_compare_end+. These methods can be implemented to execute custom code before and after the comparison against the reference distribution is made.

In the case of streaming detectors, similar methods are provided for the \verb+update+ method. All \textit{callbacks} detectors share \verb+on_fit_start+ and \verb+on_fit_end+ provided by \verb+BaseCallback+, due to the fact that \textit{data drift} detectors can be categorised as batch or streaming.

Some examples of \textit{callbacks} already implemented and ready to use in the library include tracking the history of a \textit{concept drift} detector, computing p-values through the use of permutation tests in distance-based \textit{data drift} detectors, among others.

\section{Development}
\label{sec:development}

With the intention of following a set of open-source software development standards that allow the maintainability and extensibility of the library over time, we emphasize the following areas:

\paragraph{Continuous integration.} A continuous integration workflow based on \verb+GitHub+ \verb+Actions+ ensures that new modifications easily integrate with the existing code base and that they are compatible with multiple Python versions.

\paragraph{Documentation.} An API documentation is provided using \verb+sphinx+ and hosted on the \verb+Read+ \verb+the+ \verb+Docs+\footnote{\url{https://frouros.readthedocs.io}} website. The documentation includes some basic and advanced examples on the use of these detection methods.

\paragraph{Quality code.} To ensure minimum standards in terms of code quality, code coverage is set to be greater than 90\%, and some Python quality and style tools that comply with PEP8 standards \citep{van2001pep} are used, such as \verb+flake8+, \verb+pylint+, \verb+black+, and \verb+mypy+. 

\paragraph{Open-source.} In addition to the source code being available on \verb+GitHub+\footnote{\url{https://github.com/IFCA/frouros}}, \verb+Frouros+ package can be installed through the Python Package Index (PyPI)\footnote{\url{https://pypi.org/project/frouros}}. In terms of licensing, it is distributed under the BSD-3-Clause license.

\section{Comparison to related software}
\label{sec:comparison}

With regard to the \textit{concept drift} detection part, \verb+MOA+ \citep{bifet2010moa} has most of the methods that we are including, but they are implemented in Java. In Python, \verb+River+ \citep{JMLR:v22:20-1380}, which is focused on online machine learning and streaming data, offers some \textit{concept drift} detection methods, but only a subset of those presented here. Another Python library that contains this type of detectors is \verb+scikit-multiflow+ \citep{JMLR:v19:18-251}, but it has not been taken into account due to the fact that it was merged with the online machine learning library \verb+Creme+ \citep{creme}, resulting in the aforementioned \verb+River+ library.

For the \textit{data drift} section, \verb+Alibi Detect+ \citep{alibi-detect} has several algorithms related to the field of statistical two-sample hypothesis testing, and some of them can act both online (single sample) and offline (batch sample). \verb+TorchDrift+ \cite{torchdrift} also implements some statistical two-sample hypothesis testing methods, but in this case, it uses \verb+PyTorch+ \citep{paszke2019pytorch} for their implementation.

To the best of our knowledge, \verb+Menelaus+ \cite{menelaus} is the only open-source library that has both \textit{concept} and \textit{data drift} methods, although they classify them into the following types: \textit{change detection}, \textit{concept drift}, and \textit{data drift}. \textit{Concept drift} methods are implemented in such a way that the user must necessarily be in charge of controlling each iteration of the sample without offering some helper functions or classes to interact with the detector, as \verb+Frouros+ does with the \textit{callbacks} concept that is explained in Section~\ref{subsec:callbacks}.

Table \ref{tab:frameworks-algorithms} provides a more detailed view of the methods implemented in each of the libraries mentioned above, as well as those included in \verb+Frouros+. At the time of writing this paper, \verb+Frouros+ is listed as the library with the highest number of available methods with 28.

Moreover, several other libraries and tools have been excluded from Table \ref{tab:frameworks-algorithms} due to the fact that they implement a more limited number of methods or are more focused on building graphical dashboards and visual representations, such as \verb+Deepchecks+ \citep{deepchecks}, \verb+Eurybia+ \citep{eurybia}, \verb+Evidently+ \citep{evidently}, \verb+NannyML+ \citep{nannyml}, or \verb+UpTrain+ \citep{uptrain}.

\begin{table}[htbp]
\centering
\resizebox{0.84\linewidth}{!} {
\begin{tabular}{c|c|cccccc|c|}
\cline{2-9}
 & \textbf{Method} & {\rotatebox{90}{\textbf{Alibi-detect}}} & \rotatebox{90}{\textbf{Menelaus}} & \rotatebox{90}{\textbf{MOA}} & \rotatebox{90}{\textbf{River}} & \rotatebox{90}{\textbf{TorchDrift}} & \rotatebox{90}{\textbf{Frouros}} & \textbf{Reference} \\
\hline
\multicolumn{1}{|c|}{\multirow{18}{*}{\rotatebox[origin=c]{90}{\textbf{Concept drift}}}} & ADWIN & & \checkmark & \checkmark & \checkmark & & \checkmark & \cite{bifet2007learning} \\
\multicolumn{1}{|c|}{} & BOCD & & & & & & \checkmark & \cite{adams2007bayesian} \\
\multicolumn{1}{|c|}{} & CUSUM & & \checkmark & \checkmark & & & \checkmark & \cite{page1954continuous} \\
\multicolumn{1}{|c|}{} & DDM & & \checkmark & \checkmark & \checkmark & & \checkmark & \cite{gama2004learning} \\
\multicolumn{1}{|c|}{} & ECDD-WT & & & \checkmark & & & \checkmark & \cite{ross2012exponentially} \\
\multicolumn{1}{|c|}{} & EDDM & & \checkmark & \checkmark & \checkmark & & \checkmark & \cite{baena2006early} \\
\multicolumn{1}{|c|}{} & GMA & & & \checkmark & & & \checkmark & \cite{10.2307/1266443} \\
\multicolumn{1}{|c|}{} & HDDM-A & & & \checkmark & \checkmark & & \checkmark & \cite{frias2014online} \\
\multicolumn{1}{|c|}{} & HDDM-W & & & \checkmark & \checkmark & & \checkmark & \cite{frias2014online} \\
\multicolumn{1}{|c|}{} & KSWIN & & & & \checkmark & & \checkmark & \cite{raab2020reactive} \\
\multicolumn{1}{|c|}{} & LFR & & \checkmark & & & & & \cite{wang2015concept} \\
\multicolumn{1}{|c|}{} & Page Hinkley & & \checkmark & \checkmark & \checkmark & & \checkmark & \cite{page1954continuous} \\
\multicolumn{1}{|c|}{} & RDDM & & & \checkmark & & & \checkmark & \cite{barros2017rddm} \\
\multicolumn{1}{|c|}{} & SEED & & & \checkmark & & & & \cite{huang2014detecting} \\
\multicolumn{1}{|c|}{} & SeqDrift1 & & & \checkmark & & & & \cite{sakthithasan2013one} \\
\multicolumn{1}{|c|}{} & SeqDrift2 & & & \checkmark & & & & \cite{pears2014detecting} \\
\multicolumn{1}{|c|}{} & STEPD & & \checkmark & \checkmark & & & \checkmark & \cite{nishida2007detecting} \\
\multicolumn{1}{|c|}{} & MD3 & & \checkmark & & & & & \cite{SETHI201777} \\
\hline
\multicolumn{1}{|c|}{\multirow{25}{*}{\rotatebox[origin=c]{90}{\textbf{Data drift}}} } & Anderson-Darling test & & & & & & \checkmark & \cite{scholz1987k} \\
\multicolumn{1}{|c|}{} & Bhattacharyya Distance & & & & & & \checkmark & \cite{bhattacharyya1946measure} \\
\multicolumn{1}{|c|}{} & C2ST & \checkmark & & & & & & \cite{lopez2016revisiting} \\
\multicolumn{1}{|c|}{} & CDBD & & \checkmark & & & & & \cite{lindstrom2013drift} \\
\multicolumn{1}{|c|}{} & Context-aware MMD & \checkmark & & & & & & \cite{cobb2022context} \\
\multicolumn{1}{|c|}{} & Cramér-von Mises test  & \checkmark & & & & & \checkmark & \cite{cramer1928composition} \\
\multicolumn{1}{|c|}{} & Earth Mover's Distance & & & & & \checkmark & \checkmark & \cite{rubner2000earth} \\
\multicolumn{1}{|c|}{} & Fisher’s Exact test & \checkmark & & & & & & \cite{upton1992fisher} \\
\multicolumn{1}{|c|}{} & Hellinger Distance & & & & & & \checkmark & \cite{hellinger1909neue} \\
\multicolumn{1}{|c|}{} & HDDDM & & \checkmark & & & & & \cite{ditzler2011hellinger} \\
\multicolumn{1}{|c|}{} & Histogram Intersection & & & & & & \checkmark & \cite{swain1991color} \\
\multicolumn{1}{|c|}{} & Incremental KS test & & & & & & \checkmark & \cite{dos2016fast} \\
\multicolumn{1}{|c|}{} & JS Divergence & & & & & & \checkmark & \cite{lin1991divergence} \\
\multicolumn{1}{|c|}{} & kdq-Tree & & \checkmark & & & & & \cite{dasu2006information} \\
\multicolumn{1}{|c|}{} & KL Divergence & & & & & & \checkmark & \cite{kullback1951information} \\
\multicolumn{1}{|c|}{} & Kolmogorov-Smirnov test & \checkmark & & & & \checkmark & \checkmark & \cite{massey1951kolmogorov} \\
\multicolumn{1}{|c|}{} & Learned Kernel MMD & \checkmark & & & & & & \cite{liu2020learning} \\
\multicolumn{1}{|c|}{} & LSDD & \checkmark & & & & & & \cite{bu2016pdf} \\
\multicolumn{1}{|c|}{} & Mann-Whitney U test & & & & & & \checkmark & \cite{mann1947test} \\
\multicolumn{1}{|c|}{} & MMD & \checkmark & & & & \checkmark & \checkmark & \cite{JMLR:v13:gretton12a} \\
\multicolumn{1}{|c|}{} & NN-DVI & & \checkmark & & & & & \cite{liu2018accumulating} \\
\multicolumn{1}{|c|}{} & Partial MMD & & & & & \checkmark & & \cite{viehmann2021partial} \\
\multicolumn{1}{|c|}{} & PCA-CD & & \checkmark & & & & & \cite{qahtan2015pca} \\
\multicolumn{1}{|c|}{} & PSI & & & & & & \checkmark & \cite{wu2010enterprise} \\
\multicolumn{1}{|c|}{} & Welch's t-test & & & & & & \checkmark & \cite{welch1947generalization} \\
\multicolumn{1}{|c|}{} & ${\chi}^2$ & \checkmark & & & & & \checkmark & \cite{pearson1900x} \\
\hline
& {\textbf{\# Methods}} & 9 & 13 & 14 & 7 & 4 & \textbf{28} & \\
\cline{2-9}
\end{tabular}}
\caption{Drift detection methods by library}
\label{tab:frameworks-algorithms}
\end{table}

\section{Conclusion and future work}
\label{sec:conclusion}

This paper presents \verb+Frouros+, an open-source Python library for drift detection in machine learning systems that can be used with any machine learning framework, both for methods that aim to detect \textit{concept drift} and those that try to detect \textit{data drift}. Moreover, this library provides enough flexibility to adapt to the user's workflow through the so-called \textit{callbacks} concept, which is also used in other well-known machine learning libraries. Additionally, \verb+Frouros+ aims to meet open-source software development standards that allow for easy extension, both in terms of adding new methods or \textit{callbacks}.

For future work, in addition to regularly adding new detection methods, we plan to improve the performance of \textit{concept drift} methods using Python code optimizers like Cython \cite{behnel2010cython} and adapt \textit{data drift} methods to be used on GPUs. Furthermore, adding new \textit{callbacks}, as described in Section \ref{subsec:callbacks}, would enable the library to handle even more real-world use cases.


\acks{The authors acknowledge the funding from the Agencia Estatal de Investigación, Unidad de Excelencia María de Maeztu, ref. MDM-2017-0765 and the support from the project AI4EOSC “Artificial Intelligence for the European Open Science Cloud” that has
received funding from the European Union’s Horizon Europe Research and Innovation Programme under grant agreement 101058593.}


\newpage
\bibliography{bibliography}

\end{document}